%% file: acl2019.tex
\documentclass[11pt,a4paper]{article}
\usepackage[hyperref]{acl2019}
\usepackage{times}
\usepackage{latexsym}

\usepackage{multirow}
\usepackage{amsmath}
\usepackage{amsfonts}
\usepackage{graphicx}
\usepackage{amssymb}
\usepackage{epsfig}
\usepackage{booktabs}
\usepackage{url}

\graphicspath{ {fig/} }
\aclfinalcopy 

\newcommand{\dset}{Image Editing Request }

\title{Expressing Visual Relationships via Language}

\author{
Hao Tan$^{1,2}$, Franck Dernoncourt$^{2}$, Zhe Lin$^{2}$, Trung Bui$^{2}$, Mohit Bansal$^{1}$ \\
$^{1}$UNC Chapel Hill~~~~~$^{2}$Adobe Research \\
\small\texttt{ \{haotan, mbansal\}@cs.unc.edu, \{dernonco, zlin, bui\}@adobe.com}
}

\date{}

\begin{document}
\maketitle
\begin{abstract}
Describing images with text is a fundamental problem in vision-language research. Current studies in this domain mostly focus on single image captioning. However, in various real applications (e.g., image editing, difference interpretation, and retrieval), generating relational captions for two images, can also be very useful. This important problem has not been explored mostly due to lack of datasets and effective models. To push forward the research in this direction, we first introduce a new language-guided image editing dataset that contains a large number of real image pairs with corresponding editing instructions. We then propose a new relational speaker model based on an encoder-decoder architecture with static relational attention and sequential multi-head attention. We also extend the model with dynamic relational attention, which calculates visual alignment while decoding. Our models are evaluated on our newly collected and two public datasets consisting of image pairs annotated with relationship sentences. Experimental results, based on both automatic and human evaluation, demonstrate that our model outperforms all baselines and existing methods on all the datasets.\footnote{Our data and code are publicly available at: \\ \url{https://github.com/airsplay/VisualRelationships}}

\end{abstract}
\input{1_intro.tex}

\input{2_data.tex}

\input{3_model.tex}
\input{4_result.tex}
\input{5_related.tex}

\vspace{-4pt}
\section{Conclusion}
\vspace{-4pt}
In this paper, we explored the task of describing the visual relationship between two images.
We collected the Image Editing Request dataset, which contains image pairs and human annotated editing instructions.
We designed novel relational speaker models and evaluate them on our collected and other public existing dataset.
Based on automatic and human evaluations, our relational speaker model improves the ability to capture visual relationships.
For future work, we are going to further explore the possibility to merge the three datasets by either learning a joint image representation or by transferring domain-specific knowledge.
We are also aiming to enlarge our Image Editing Request dataset with newly-released posts on Reddit and Zhopped.

\section*{Acknowledgments}
We thank the reviewers for their helpful comments and Nham Le for helping with the initial data collection. This work was supported by Adobe, ARO-YIP Award \#W911NF-18-1-0336, and faculty awards from Google, Facebook, and Salesforce. The views, opinions, and/or findings contained in this article are those of the authors and should not be interpreted as representing the official views or policies, either expressed or implied, of the funding agency.

\bibliography{acl2019}
\bibliographystyle{acl_natbib}

\end{document}

%% file: 1_intro.tex
\begin{figure}[t]
\centering
\includegraphics[width=0.45\textwidth]{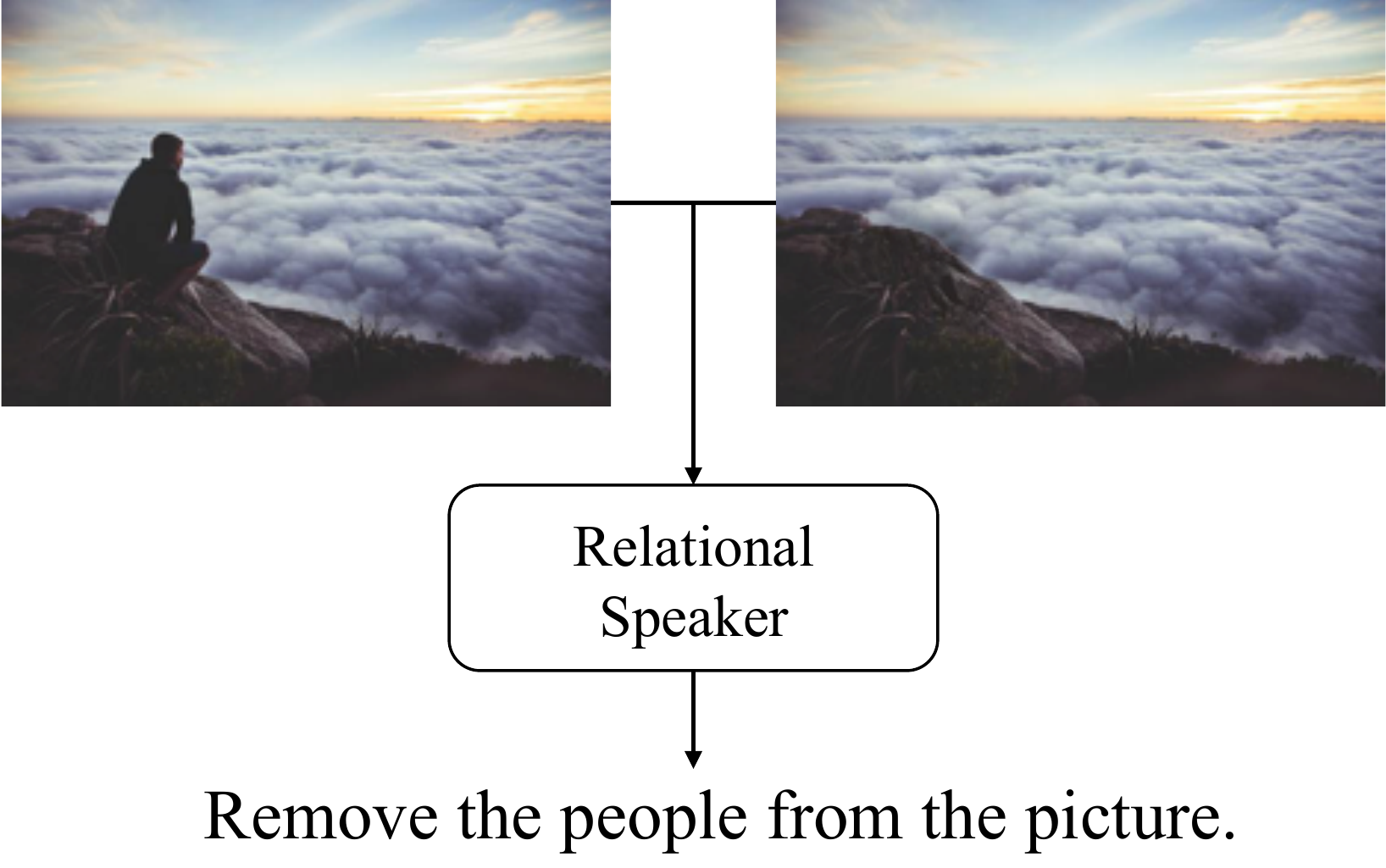}
\caption{
An example result of our method showing the input image pair from our Image Editing Request dataset, and the output instruction predicted by our relational speaker model trained on the dataset.
} 
\label{fig:intro}
\vspace{-7pt}
\end{figure}

\section{Introduction}

Generating captions to describe natural images is a fundamental research problem at the intersection of computer vision and natural language processing.
Single image captioning~\cite{mori1999image, farhadi2010every, kulkarni2011baby} has many practical applications such as text-based image search, photo curation, assisting of visually-impaired people, image understanding in social media, etc. This task has drawn significant attention in the research community with numerous studies~\cite{vinyals2015show, xu2015show, anderson2018bottom}, and recent state of the art methods have achieved promising results on large captioning datasets, such as MS COCO~\cite{lin2014microsoft}.
Besides single image captioning, the community has also explored other visual captioning problems such as video captioning~\cite{venugopalan2015sequence,xu2016msr}, and referring expressions~\cite{kazemzadeh2014referitgame, yu2017joint}.
However, the problem of two-image captioning, especially the task of describing the relationships and differences between two images, is still under-explored. In this paper, we focus on advancing research in this challenging problem by introducing a new dataset and proposing novel neural relational-speaker models.

To the best of our knowledge, \newcite{jhamtani2018learning} is the only public dataset aimed at generating natural language descriptions for two real images. 
This dataset is about `spotting the difference', and hence focuses more on describing exhaustive differences by learning alignments between multiple text descriptions and multiple image regions; 
hence the differences are intended to be explicitly identifiable by subtracting two images.
There are many other tasks that require more diverse, detailed and implicit relationships between two images.
Interpreting image editing effects with instructions is a suitable task for this purpose, because it has requirements of exploiting visual transformations and it is widely used in real life, such as explanation of complex image editing effects for laypersons or visually-impaired users, image edit or tutorial retrieval, and language-guided image editing systems.
We first build a new language-guided image editing dataset with high quality annotations by (1) crawling image pairs from real image editing request websites, (2) annotating editing instructions via Amazon Mechanical Turk, and (3) refining the annotations through experts.

Next, we propose a new neural speaker model for generating sentences that describe the visual relationship between a pair of images. Our model is general and not dependent on any specific dataset.
Starting from an attentive encoder-decoder baseline, we first develop a model enhanced with two attention-based neural components, a static relational attention and a sequential multi-head attention, to address these two challenges, respectively.
We further extend it by designing a dynamic relational attention module to combine the advantages of these two components, which finds the relationship between two images while decoding.
The computation of dynamic relational attention is mathematically equivalent to attention over all visual ``relationships''.
Thus, our method provides a direct way to model visual relationships in language. 

To show the effectiveness of our models, we evaluate them on three datasets: our new dataset, the "Spot-the-Diff" dataset~\cite{jhamtani2018learning}, and the two-image visual reasoning NLVR2 dataset~\cite{suhr2018corpus} (adapted for our task). 
We train models separately on each dataset with the same hyper-parameters and evaluate them on the same test set across all methods.
Experimental results demonstrate that our model outperforms  all the baselines and existing methods.
The main contributions of our paper are: 
(1) We create a novel human language guided image editing dataset to boost the study in describing visual relationships;
(2) We design novel relational-speaker models, including a dynamic relational attention module, to handle the problem of two-image captioning by focusing on all their visual relationships;
(3) Our method is evaluated on several datasets and achieves the state-of-the-art.

%% file: 2_data.tex
\begin{figure*}[t]
\centering
\includegraphics[width=0.95\textwidth]{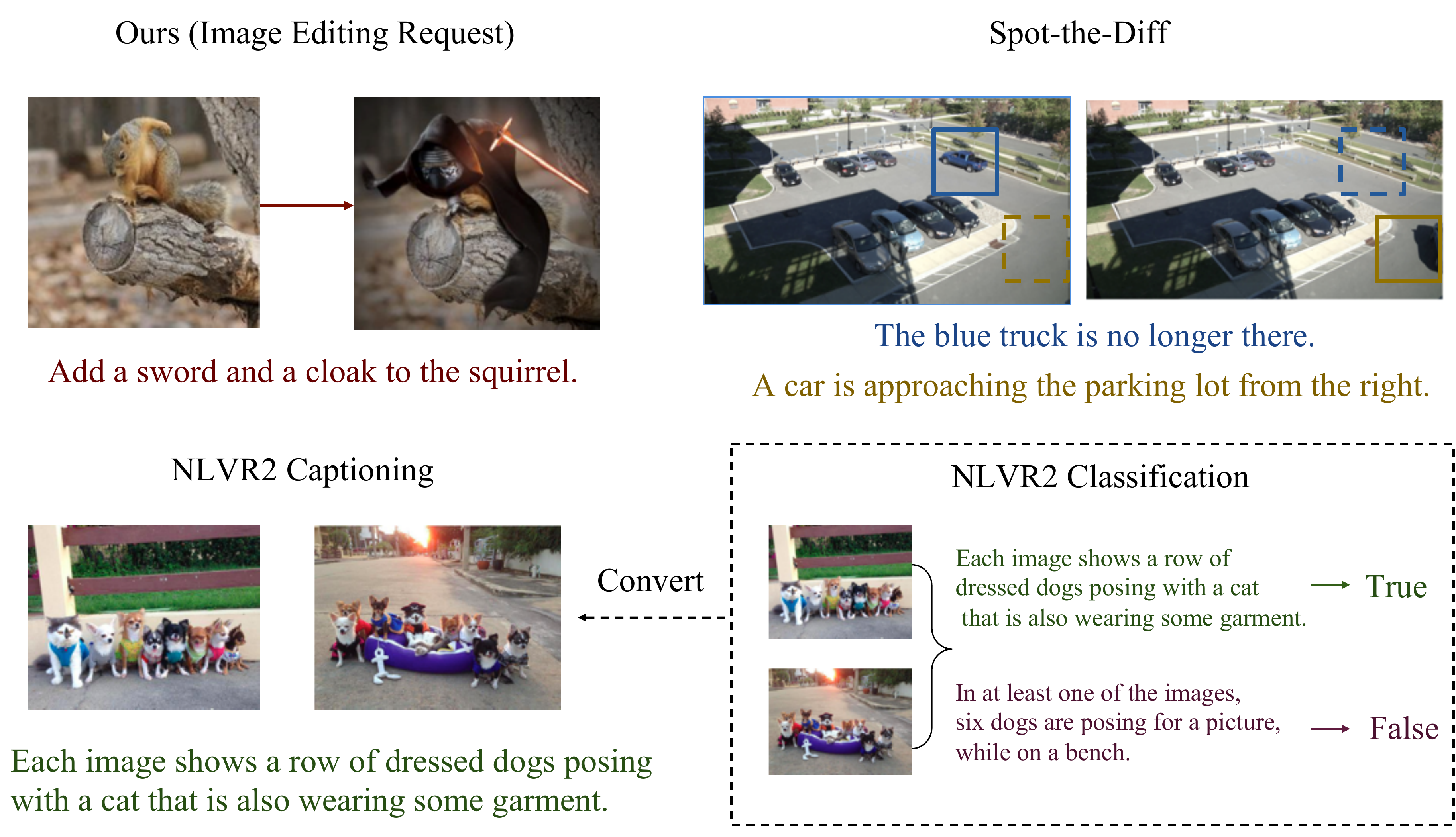}
\caption{Examples from three datasets: our Image Editing Request, Spot-the-Diff, and NLVR2. 
Each example involves two natural images and an associated sentence describing their relationship.
The task of generating NLVR2 captions is converted from its original classification task.}
\label{fig:datasets}

\end{figure*}

\section{Datasets}
\label{sec:dataset}
We present the collection process and statistics of our \dset dataset and briefly introduce two public datasets (viz., Spot-the-Diff and NLVR2). All three datasets are used to study the task of two-image captioning and evaluating our relational-speaker models.
Examples from these three datasets are shown in Fig.~\ref{fig:datasets}. 
\subsection{\dset  Dataset}
\label{sec:our_dataset}
Each instance in our dataset consists of an image pair (i.e., a source image and a target image) and a corresponding editing instruction which correctly and comprehensively describes the transformation from the source image to the target image.
Our collected \dset dataset will be publicly released along with the scripts to unify it with the other two datasets.
\subsubsection{Collection Process}
To create a high-quality, diverse dataset, we follow a three-step pipeline: image pairs collection, editing instructions annotation, and post-processing by experts (i.e., cleaning and test set annotations labeling).
\paragraph{Images Pairs Collection}
We first crawl the editing image pairs (i.e., a source image and a target image) from posts on Reddit (Photoshop request subreddit)\footnote{https://www.reddit.com/r/photoshoprequest} and Zhopped\footnote{http://zhopped.com}. 
Posts generally start with an original image and an editing specification.
Other users would send their modified images by replying to the posts.
We collect original images and modified images as source images and target images, respectively. 
\paragraph{Editing Instruction Annotation}
The texts in the original Reddit and Zhopped posts are too noisy to be used as image editing instructions. 
To address this problem, 
we collect the image editing instructions on MTurk using an
interactive interface 
that allows the MTurk annotators to either write an image editing instruction corresponding to a displayed image pair, or flag it as invalid (e.g., if the two images have nothing in common).
\begin{table}[]
\centering
\small
\begin{tabular}{|c|c|c|c|c|c|}
\hline
              & B-1 & B-2 & B-3 & B-4 & Rouge-L \\ \hline \hline
Ours          & 52     & 34     & 21     & 13     & 45      \\ 
Spot-the-Diff & 41     & 25     & 15     & 8      & 31      \\ 
MS COCO       & 38     & 22     & 15     & 8      & 34      \\ \hline
\end{tabular}
\caption{Human agreement on our datasets, compared with Spot-the-Diff and MS COCO (captions=3). B-1 to B-4 are BLEU-1 to BLEU-4. Our dataset has the highest human agreement. }
\label{table:human_agree}
\end{table}
\paragraph{Post-Processing by Experts}
Mturk annotators are not always experts in image editing.
To ensure the quality of the dataset, we hire an image editing expert to label each image editing instruction of the dataset as one of the following four options: 1. correct instruction, 2. incomplete instruction, 3. implicit request, 4. other type of errors. 
Only the data instances labeled with ``correct instruction'' are selected to compose our dataset, and are used in training or evaluating our neural speaker model.

Moreover, two additional experts are required to write two more editing instructions (one instruction per expert) for each image pair in the validation and test sets.
This process enables the dataset to be a multi-reference one, which allows various automatic evaluation metrics, such as BLEU, CIDEr, and ROUGE to more accurately evaluate the quality of generated sentences.

\subsubsection{Dataset Statistics}
\begin{figure}[t]
\centering
\vspace{-8pt}
\includegraphics[width=0.35\textwidth]{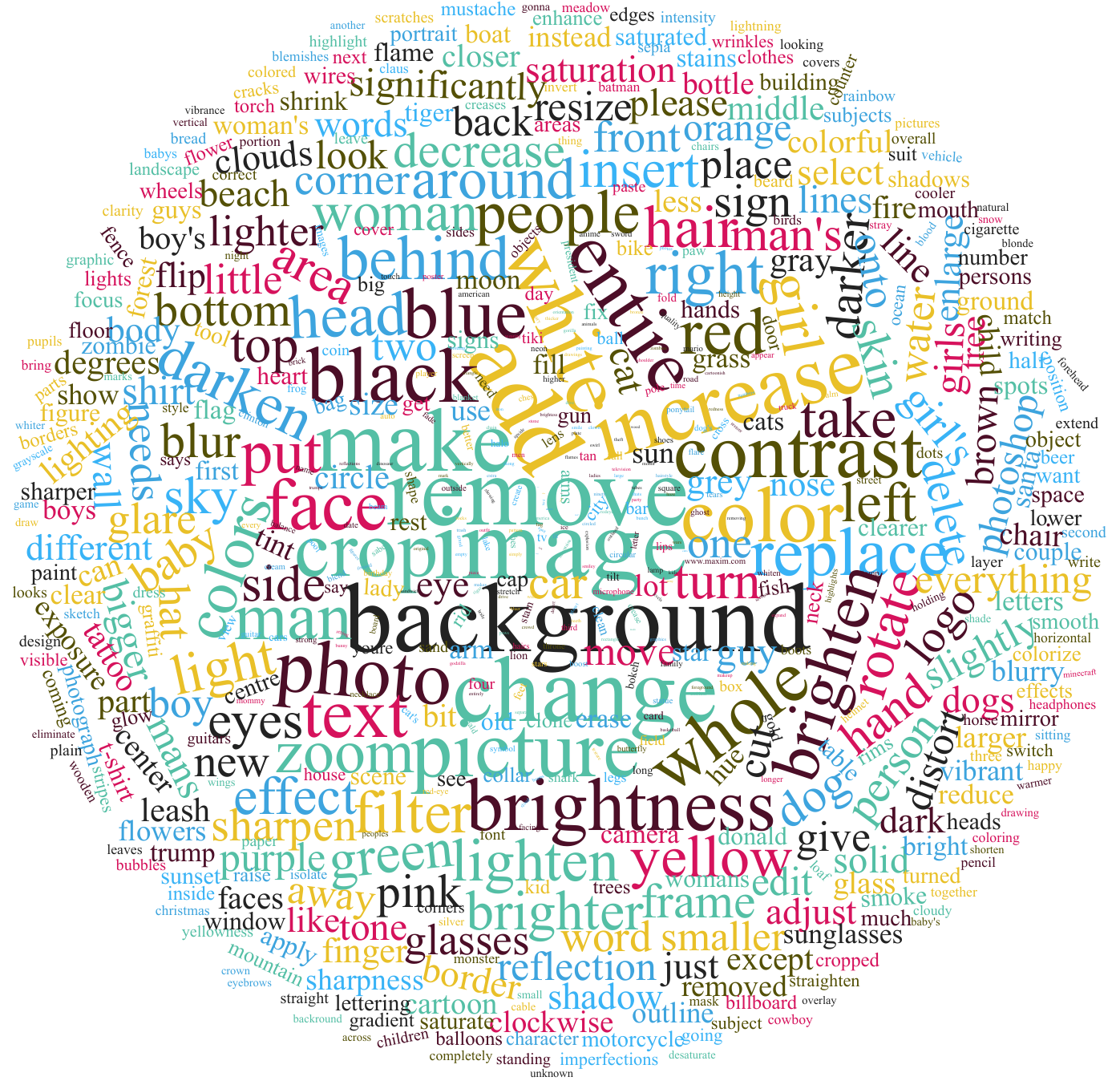}
\caption{
Word cloud showing the vocabulary frequencies of our \dset dataset.}   
\label{fig:words}
\end{figure}

The \dset dataset that we have collected and annotated currently contains $3\mbox{,}939$ image pairs ($3061$ in training, $383$ in validation, $495$ in test) with $5\mbox{,}695$ human-annotated instructions in total.
Each image pair in the training set has one instruction, and each image pair in the validation and test sets has three instructions, written by three different annotators.
Instructions have an average length of $7.5$ words (standard deviation: $4.8$).
After removing the words with less than three occurrences, the dataset has a vocabulary of $786$ words. 
The human agreement of our dataset is shown in Table~\ref{table:human_agree}. 
The word frequencies in our dataset are visualized in Fig.~\ref{fig:words}.
Most of the images in our dataset are realistic. Since the task is image editing, target images may have some artifacts (see  Image Editing Request examples in Fig.~\ref{fig:datasets} and Fig.~\ref{fig:example}).

\begin{figure*}[t]
\centering
\includegraphics[width=0.98\textwidth]{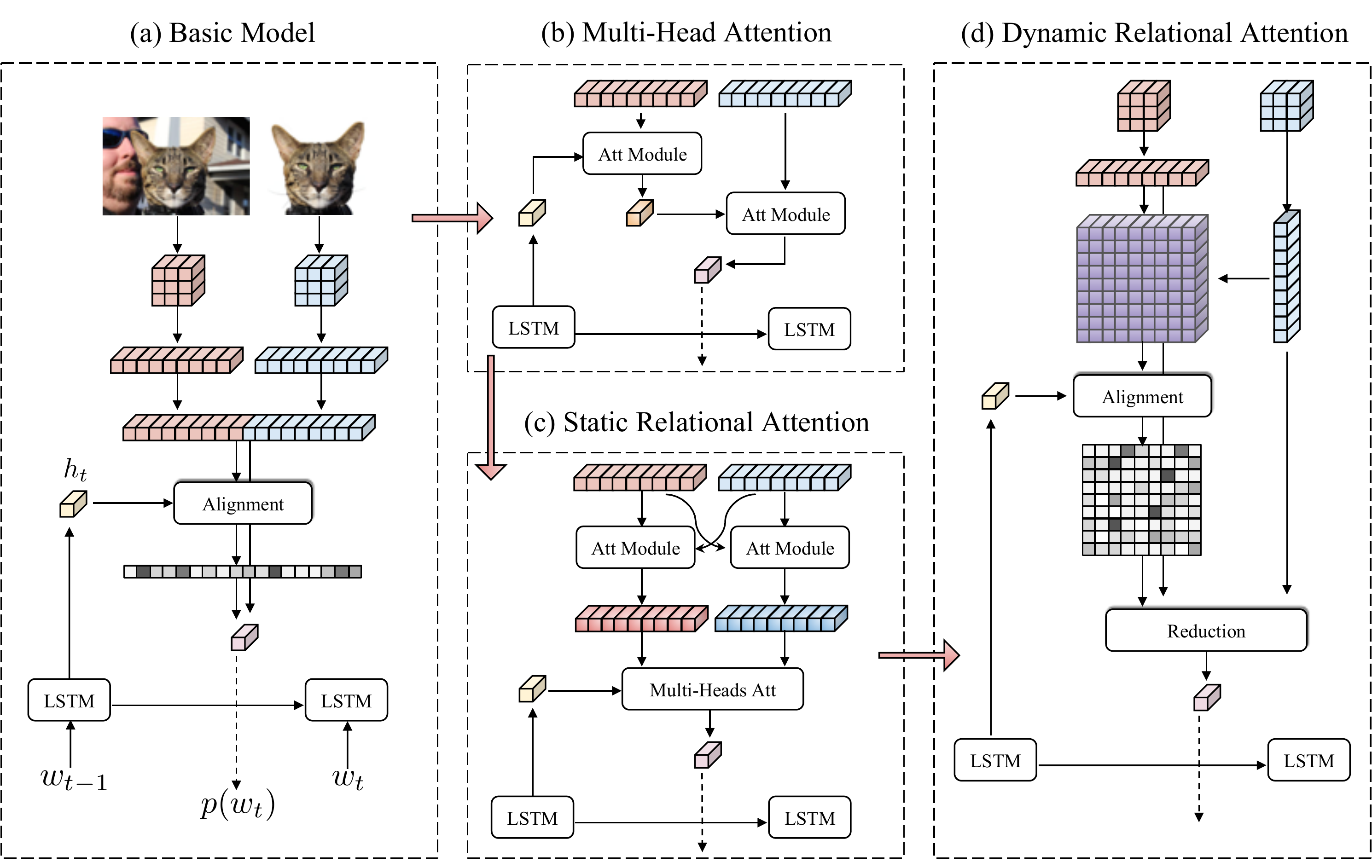}
\caption{The evolution diagram of our models to describe the visual relationships. 
One decoding step at $t$ is shown.
The linear layers are omitted for clarity. 
The basic model (a) is an attentive encoder-decoder model, which is enhanced by the multi-head attention (b) and static relational attention (c). Our best model (d) dynamically computes the relational scores in decoding to avoid losing relationship information.
}
\label{fig:model}
\end{figure*}

\subsection{Existing Public Datasets}
\label{sec:other_datasets}
To show the generalization of our speaker model, we also train and evaluate our model on two public datasets,  Spot-the-Diff~\cite{jhamtani2018learning} and NLVR2~\cite{suhr2018corpus}.
Instances in these two datasets are each composed of two natural images and a human written sentence describing the relationship between the two images. To the best of our knowledge, these are the only two public datasets with a reasonable amount of data that are suitable for our task.
We next briefly introduce these two datasets.
\paragraph{Spot-the-Diff}
This dataset is designed to help generate a set of instructions that can comprehensively describe all visual differences.
Thus, the dataset contains images from video-surveillance footage, in which differences can be easily found. This is because all the differences could be effectively captured by subtractions between two images, as shown in Fig.~\ref{fig:datasets}.
The dataset contains $13\mbox{,}192$ image pairs, and an average of $1.86$ captions are collected for each image pair.
The dataset is split into training, validation, and test sets with a ratio of $8\mbox{:}1\mbox{:}1$.
\paragraph{NLVR2}
The original task of Cornell Natural Language for Visual Reasoning (NLVR2) dataset is visual sentence classification, see Fig.~\ref{fig:datasets} for an example.
Given two related images and a natural language statement as inputs, a learned model needs to determine whether the statement correctly describes the visual contents.
We convert this classification task to a generation task by taking only the image pairs with correct descriptions.
After conversion, the amount of data is $51\mbox{,}020$, which is almost half of the original dataset with a size of $107\mbox{,}296$.
We also preserve the training, validation, and test split in the original dataset.

%% file: 3_model.tex
\section{Relational Speaker Models}
\label{sec:model}
In this section, we aim to design a general speaker model that describes the relationship between two images.
Due to the different kinds of visual relationships, the meanings of images vary in different tasks: ``before'' and ``after'' in Spot-the-Diff, ``left'' and ``right'' in NLVR2, ``source'' and ``target'' in our \dset dataset. 
We use the nomenclature of ``source'' and ``target'' for simplification, but our model is general and not designed for any specific dataset.
Formally, the model generates a sentence \{$w_1, w_2, ..., w_{T}$\} describing the relationship between the source image $I^\textsc{src}$ and the target image $I^\textsc{trg}$. $\{w_t\}_{t=1}^T$ are the word tokens with a total length of $T$. $I^\textsc{src}$ and $I^\textsc{trg}$ are natural images in their raw RGB pixels. In the rest of this section, we first introduce our basic attentive encoder-decoder model, and show how we gradually improve it to fit the task better.
\subsection{Basic Model}
\label{sec:basic_model}
Our basic model (Fig.~\ref{fig:model}(a)) is similar to the baseline model in \newcite{jhamtani2018learning}, which is adapted from the attentive encoder-decoder model for single image captioning~\cite{xu2015show}. 
We use ResNet-$101$~\cite{he2016deep} as the feature extractor to encode the source image $I^\textsc{src}$ and the target image $I^\textsc{trg}$. 
The feature maps of size $N\times N \times 2048$ are extracted, where $N$ is the height or width of the feature map.
Each feature in the feature map represents a part of the image. 
Feature maps are then flattened to two $N^2 \times 2048$ feature sequences $f^\textsc{src}$ and $f^\textsc{trg}$, which are further concatenated to a single feature sequence $f$.
\vspace{-7pt}
\begin{align}
    f^\textsc{src} &  = \mathrm{ResNet}\left(I^\textsc{src}\right) \\
    f^\textsc{trg} & = \mathrm{ResNet}\left(I^\textsc{trg}\right) \\
    f & = \left[ f^\textsc{src}_1, \ldots, f^\textsc{src}_{N^2}, f^\textsc{trg}_1,
          , \ldots, f^\textsc{trg}_{N^2} \right]
          \vspace{-10pt}
\end{align}
At each decoding step $t$, the LSTM cell takes the embedding of the previous word $w_{t-1}$ as an input.
The word $w_{t-1}$ either comes from the ground truth (in training) or takes the token with maximal probability (in evaluating).
The attention module then attends to the feature sequence $f$ with the hidden output $h_t$ as a query.
Inside the attention module, it first computes the alignment scores $\alpha_{t,i}$ between the query $h_t$ and each $f_i$.
Next, the feature sequence $f$ is aggregated with a weighted average (with a weight of $\alpha$) to form the image context $\hat{f}$.
Lastly, the context $\hat{f}_t$ and the hidden vector $h_t$ are merged into an attentive hidden vector $\hat{h}_t$ with a fully-connected layer:
\begin{align}
    \tilde{w}_{t-1} & = \mathrm{embedding} \left(w_{t-1} \right) \\
    h_t, c_t & = \mathrm{LSTM} \left(\tilde{w}_{t-1},  h_{t-1}, c_{t-1} \right) \\
    \alpha_{t,i} & = \mathrm{softmax}_i \left(h_t^\top W_\textsc{img} f_i \right) \\
    \hat{f}_t & = \sum_i \alpha_{t,i} f_i \\
    \hat{h}_t & = \tanh( W_1 [\hat{f}_t; h_t] + b_1)
\end{align}
The probability of generating the $k$-th word token at time step $t$ is softmax over a linear transformation of the attentive hidden $\hat{h}_t$.
The loss $\mathcal{L}_t$ is the negative log likelihood of the ground truth word token $w^*_t$:
\begin{align}
    p_t(w_{t,k}) & = \mathrm{softmax}_k \left(W_\textsc{w}\, \hat{h}_t + b_\textsc{w}\right) \\
    \mathcal{L}_t & = - \log p_t(w^*_{t})
\end{align}

\subsection{Sequential Multi-Head Attention}
One weakness of the basic model is that the plain attention module simply takes the concatenated image feature $f$ as the input, which does not differentiate between the two images.
We thus consider applying a multi-head attention module~\cite{vaswani2017attention} to handle this (Fig.~\ref{fig:model}(b)). 
Instead of using the simultaneous multi-head attention~\footnote{We also tried the original multi-head attention but it is empirically weaker than our sequential multi-head attention.} in Transformer~\cite{vaswani2017attention}, we implement the multi-head attention in a sequential way. This way, when the model is attending to the target image, the contextual information retrieved from the source image is available and  can therefore perform better at differentiation or relationship learning.

In detail, the source attention head first attends to the flattened source image feature $f^\textsc{src}$. 
The attention module is built in the same way as in Sec.~\ref{sec:basic_model}, except that it now only attends to the source image:
\begin{align}
    \alpha^\textsc{src}_{t,i} & = \mathrm{softmax}_i (h_t^\top W_\textsc{src} f^\textsc{src}_i) \\
    \hat{f}^\textsc{src}_t & = \sum_i \alpha^\textsc{src}_{t,i} f^\textsc{src}_i \\
    \hat{h}^\textsc{src}_t & = \tanh( W_2 [ \hat{f}^\textsc{src}_t; h_t] + b_2)
\end{align}

The target attention head then takes the output of the source attention $\hat{h}^\textsc{src}_t$ as a query to retrieve appropriate information from the target feature $f^\textsc{trg}$:
\begin{align}
    \alpha^\textsc{trg}_{t,j} & = \mathrm{softmax}_j (\hat{h}_t^{\textsc{src}\top} W_\textsc{trg} f^\textsc{trg}_j) \\
    \hat{f}^\textsc{trg}_t & = \sum_j \alpha^\textsc{trg}_{t,j} f^\textsc{trg}_j\\
    \hat{h}^\textsc{trg}_t & = \tanh( W_3 [\hat{f}^\textsc{trg}_t; \hat{h}^\textsc{src}_t] + b_3)
\end{align}
In place of $\hat{h}_t$, the output of the target head $\hat{h}^\textsc{trg}_t$ is used to predict the next word.\footnote{We tried to exchange the order of two heads or have two orders concurrently. We didn't see any significant difference in results between them.}

\begin{table*}[]
\centering
\begin{tabular}{|c|c|c|c|c|}
\hline
Method                   &   BLEU-4 &  CIDEr          &  METEOR         &  ROUGE-L \\ \hline \hline
\multicolumn{5}{|c|}{Our Dataset (Image Editing Request)}                                                       \\ \hline \hline
basic model              & 5.04   & 21.58          & 11.58          & 34.66   \\ \hline
+multi-head att          & 6.13   & 22.82          & 11.76          & 35.13   \\ \hline
+static rel-att          & 5.76   & 20.70          & 12.59          & 35.46   \\ \hline
-static +dynamic rel-att & \textbf{6.72 }  & \textbf{26.36 }         & \textbf{12.80} & \textbf{37.25 }  \\ \hline \hline
\multicolumn{5}{|c|}{Spot-the-Diff}                                                                      \\ \hline \hline
{\small CAPT}\small\cite{jhamtani2018learning}                     & 7.30   & 26.30          & 10.50          & 25.60   \\ \hline
{\small DDLA}\small\cite{jhamtani2018learning}                     & \textbf{8.50 }  & 32.80          & 12.00          & 28.60   \\ \hline
basic model              & 5.68   & 22.20          & 10.98          & 24.21   \\ \hline
+multi-head att          & 7.52   & 31.39          & 11.64          & 26.96   \\ \hline
+static rel-att          & 8.31   & 33.98          & \textbf{12.95 }         & 28.26   \\ \hline
-static +dynamic rel-att & 8.09   & \textbf{35.25} & 12.20          & \textbf{31.38 }  \\ \hline \hline
\multicolumn{5}{|c|}{NLVR2}                                                   \\ \hline \hline
basic model              & 5.04   & 43.39          & 10.82          & 22.19   \\ \hline
+multi-head att          &\textbf{ 5.11  } & 44.80          & 10.72          & 22.60   \\ \hline
+static rel-att          & 4.95   & 45.67          & \textbf{10.89 }         & 22.69   \\ \hline
-static +dynamic rel-att & 5.00   & \textbf{46.41} & 10.37          & \textbf{22.94 }  \\ \hline
\end{tabular}
\caption{
Automatic metric of test results on three datasets. Best results of the main metric are marked in bold font. 
Our full model is the best on all three datasets with the main metric.
}
\label{table:result}
\end{table*}

\subsection{Static Relational Attention}
Although the sequential multi-head attention model can learn to differentiate the two images, visual relationships are not explicitly examined.
We thus allow the model to statically (i.e., not in decoding) compute the relational score between source and target feature sequences and reduce the scores into two relationship-aware feature sequences.
We apply a bi-directional relational attention (Fig.~\ref{fig:model}(c)) for this purpose: one from the source to the target, and one from the target to the source.
For each feature in the source feature sequence, the source-to-target attention computes its alignment with the features in the target feature sequences.
The source feature, the attended target feature, and the difference between them are then merged together with a fully-connected layer:
\begin{align}
    \alpha^\textsc{s$\rightarrow$t}_{i, j} &= \mathrm{softmax}_j( (W_\textsc{s} f^\textsc{src}_i) ^\top  (W_\textsc{t} f^\textsc{trg}_j) ) \\
    \hat{f}^\textsc{s$\rightarrow$t}_i &= \sum_j \alpha^\textsc{s$\rightarrow$t}_{i, j} f^\textsc{trg}_j \\
    \hat{f}^\textsc{s}_i & = \tanh ( W_\textsc{4}[f^\textsc{src}_i ; \hat{f}^\textsc{s$\rightarrow$t}_i] + b_\textsc{4})
\end{align}
We decompose the attention weight into two small matrices $W_\textsc{s}$ and $W_\textsc{t}$ so as to reduce the number of parameters, because the dimension of the image feature is usually large. 
The target-to-source cross-attention is built in an opposite way: it takes each target feature $f^\textsc{trg}_j$ as a query, attends to the source feature sequence, and get the attentive feature $\hat{f}^\textsc{t}_j$.
We then use these two bidirectional attentive sequences $\hat{f}^\textsc{s}_i$ and $\hat{f}^\textsc{t}_j$ in the multi-head attention module (shown in previous subsection) at each decoding step. 
\subsection{Dynamic Relational Attention}
The static relational attention module compresses pairwise relationships (of size $N^4$) into two relationship-aware feature sequences (of size $2\!\times\!N^2$). 
The compression saves computational resources but has potential drawback in information loss as discussed in \newcite{bahdanau2014neural} and \newcite{xu2015show}.
In order to avoid losing information, we modify the static relational attention module to its dynamic version, which calculates the relational scores while decoding (Fig.~\ref{fig:model}(d)).

At each decoding step $t$, the dynamic relational attention calculates the alignment score $a_{t,i,j}$ between three vectors: a source feature $f^\textsc{src}_i$, a target feature $f^\textsc{trg}_j$, and the hidden state $h_t$. 
Since the dot-product used in previous attention modules does not have a direct extension for three vectors, we extend the dot product and use it to compute the three-vector alignment score.
\begin{align}
    \mathrm{dot} (x, y)  & = \sum_d x_d \,y_d = x^\top y \\
    \mathrm{dot^*} (x, y, z)  &  = \sum_d x_d \, y_d z_d = (x \odot y)^\top z 
\end{align}
\vspace{-20pt}
\begin{align}
    a_{t,i,j}  & =  \mathrm{dot^*}( W_\textsc{sk} f^\textsc{src}_i, W_\textsc{tk} f^\textsc{trg}_j, W_\textsc{hk} h_t) \\
            \label{eqn:key_equiv}
            & = (W_\textsc{sk} f^\textsc{src}_i \odot W_\textsc{tk} f^\textsc{trg}_j)^\top W_\textsc{hk} h_t
\end{align}
where $\odot$ is the element-wise multiplication.

The alignment scores (of size $N^4$) are normalized by softmax. And the attention information is fused to the attentive hidden vector $\hat{f}^\textsc{d}_t$ as previous.
\begin{align}
    \alpha_{t,i,j} & = \mathrm{softmax}_{i,j}\left( a_{t,i,j} \right)  \\
    \hat{f}^\textsc{src-d}_t & = \sum_{i,j} \alpha_{t, i, j} f^\textsc{src}_i \\
    \hat{f}^\textsc{trg-d}_t & = \sum_{i,j} \alpha_{t, i, j} f^\textsc{trg}_j \\
    \hat{f}^\textsc{d}_t & = \tanh(W_5[\hat{f}^\textsc{src-d}_t; \hat{f}^\textsc{trg-d}_t; h_t] \mbox{+} b_5) \\
            \label{eqn:value_equiv}
    & = \tanh( W_\textsc{5s} \hat{f}^\textsc{src-d}_t + W_\textsc{5t} \hat{f}^\textsc{trg-d}_t + \nonumber\\ 
    & \;\;\;\;\;\;\;\;\;\;\;\;\; W_\textsc{5h} h_t + b_5)
\end{align}
where $W_\textsc{5s}$, $W_\textsc{5t}$, $W_\textsc{5h}$ are sub-matrices of $W_5$ and $W_5 = [W_\textsc{5s}, W_\textsc{5t}, W_\textsc{5h}]$.

According to Eqn.~\ref{eqn:key_equiv} and Eqn.~\ref{eqn:value_equiv}, we find an analog in conventional attention layers with following specifications:
\begin{itemize}
    \item Query: $h_t$
	\item Key:   $W_\textsc{sk} f^\textsc{src}_i \odot W_\textsc{tk} f^\textsc{trg}_j$
	\item Value: $W_\textsc{5s} f^\textsc{src}_i + W_\textsc{5t} f^\textsc{trg}_j$
\end{itemize}
The key $W_\textsc{sk} f^\textsc{src}_i \odot W_\textsc{tk} f^\textsc{trg}_j$ and the value $W_\textsc{5s} f^\textsc{src}_i \mbox{+} W_\textsc{5t} f^\textsc{trg}_j$ can be considered as representations of the visual relationships between $f^\textsc{src}_i$ and $f^\textsc{trg}_j$.
It is a direct attention to the visual relationship between the source and target images, hence is suitable for the task of generating relationship descriptions.

%% file: 4_result.tex
\section{Results}
\label{sec:results}
To evaluate the performance of our relational speaker models (Sec.~\ref{sec:model}), we trained them on all three datasets (Sec.~\ref{sec:dataset}). 
We evaluate our models based on both automatic metrics as well as pairwise human evaluation.
We also show our generated examples for each dataset.
\subsection{Experimental Setup}
We use the same hyperparameters when applying our model to the three datasets. Dimensions of hidden vectors are $512$.
The model is optimized by Adam with a learning rate of $1e-4$.
We add dropout layers of rate $0.5$ everywhere to avoid over-fitting.
When generating instructions for evaluation, we use maximum-decoding:
the word $w_t$ generated at time step $t$ is $\arg\max_k p(w_{t,k})$.
For the Spot-the-Diff dataset, we take the ``Single sentence decoding'' experiment as in \newcite{jhamtani2018learning}.
We also try to mix the three datasets but we do not see any  improvement. 
We also try different ways to mix the three datasets but we do not see improvement. 
We first train a unified model on the union of these datasets. 
The metrics drop a lot because the tasks and language domains (e.g., the word dictionary and lengths of sentences) are different from each other.
We next only share the visual components to overcome the disagreement in language.
However, the image domain are still quite different from each other (as shown in Fig.~\ref{fig:datasets}).
Thus, we finally separately train three models on the three datasets with minimal cross-dataset modifications.

\subsection{Metric-Based Evaluation}
\label{sec:numbers}
As shown in Table~\ref{table:result}, we compare the performance of our models on all three datasets with various automated metrics.
Results on the test sets are reported.
Following the setup in \newcite{jhamtani2018learning}, we takes CIDEr~\cite{vedantam2015cider} as the main metric in evaluating the Spot-the-Diff and NLVR2 datasets.
However, CIDEr is known as its problem in up-weighting unimportant details~\cite{kilickaya2017re, liu2017improved}. 
In our dataset, we find that instructions generated from a small set of short phrases could get a  high CIDEr score.
We thus change the main metric of our dataset to  METEOR~\cite{banerjee2005meteor}, which is manually verified to be aligned with human judgment on the validation set in our dataset. 
To avoid over-fitting, the model is early-stopped based on the main metric on validation set.
We also report the BLEU-4~\cite{papineni2002bleu} and ROUGE-L~\cite{lin2004rouge} scores.

The results on various datasets shows the gradual improvement made by our novel neural components, which are designed to better describe the relationship between 2 images.
Our full model has a significant improvement in result over baseline.
The improvement on the NLVR2 dataset is limited because the comparison of two images was not forced to be considered when generating instructions.
\begin{figure*}[t]
\centering
\includegraphics[width=0.98\textwidth]{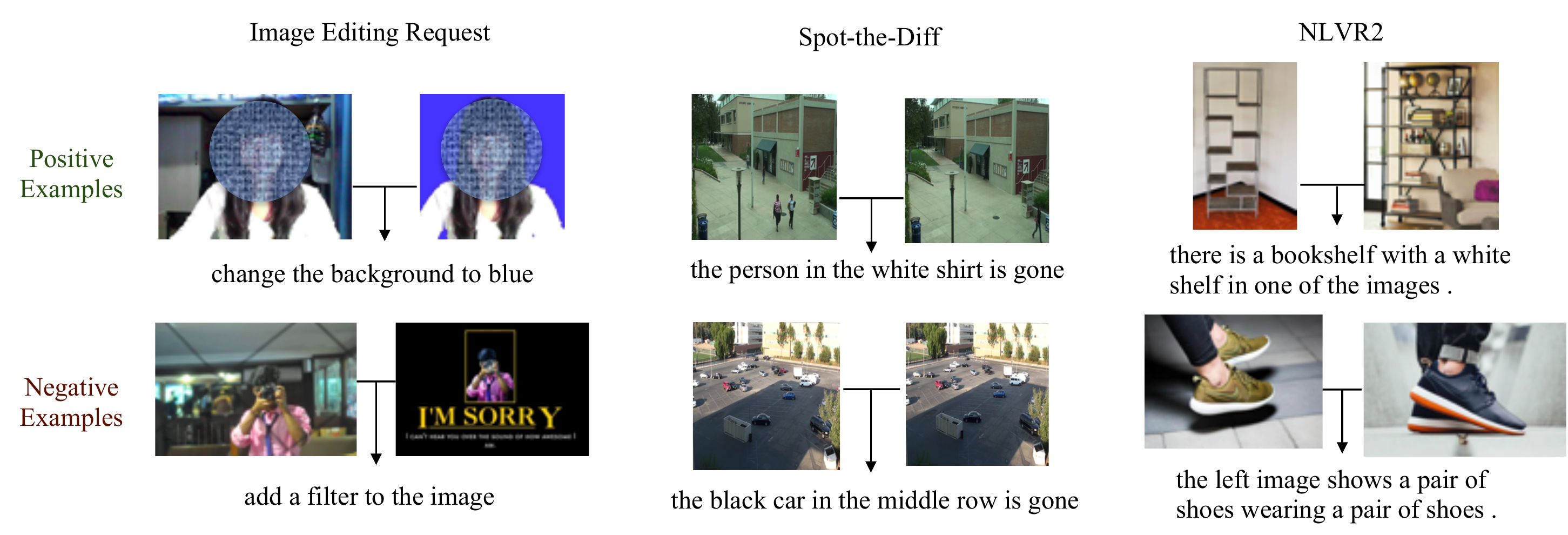}
\vspace{-7pt}
\caption{
Examples of positive and negative results of our model from the three datasets. Selfies are blurred.
}
\label{fig:example}
\vspace{-10pt}
\end{figure*}

\begin{table}[]
\small
\begin{tabular}{|c|c|c|c|c|}
\hline
              & Basic  & Full    & Both Good & Both Not \\ \hline
Ours(IEdit)          & 11         & \textbf{24} & 5         & 60       \\ \hline
Spot-the-Diff &  22          &  \textbf{37}           &    6       &  35        \\ \hline
NLVR2         & 24         & \textbf{37} & 17        & 22       \\ \hline
\end{tabular}
\caption{
Human evaluation on $100$ examples. Image pair and two captions generated by our basic model and full model are shown to the user. The user chooses one from `Basic' model wins, `Full' model wins, `Both Good', or `Both Not'. Better model marked in bold font.
}
\label{table:human_eval}
\end{table}

\subsection{Human Evaluation and Qualitative Analysis}
We conduct a pairwise human evaluation on our generated sentences, which is used in \newcite{celikyilmaz2018deep} and \newcite{pasunuru2017reinforced}.
\newcite{agarwala_2018} also shows that the pairwise comparison is better than scoring sentences individually.
We randomly select $100$ examples from the test set in each dataset and generate captions via our full speaker model.
We ask users
to choose a better instruction between the captions generated by our full model and the basic model, or alternatively indicate that the two captions are equal in quality.
The Image Editing Request dataset is specifically annotated by the image editing expert.
The winning rate of our full model (dynamic relation attention) versus the basic model is shown in Table~\ref{table:human_eval}. 
Our full model  outperforms the basic model significantly. 
We also show positive and negative examples generated by our full model in Fig.~\ref{fig:example}. 
In our Image Editing Request corpus, the model was able to detect and describe the editing actions but it failed in handling the arbitrary complex editing actions.
We keep these hard examples in our dataset to match real-world requirements and allow follow-up future works to pursue the remaining challenges in this task.
Our model is designed for non-localized relationships thus we do not explicitly model the pixel-level differences; however, we still find that the model could learn these differences in the Spot-the-Diff dataset.
Since the descriptions in Spot-the-Diff is relatively simple, the errors mostly come from wrong entities or undetected differences as shown in Fig.~\ref{fig:example}.
Our model is also sensitive to the image contents as shown in the NLVR2 dataset.

%% file: 5_related.tex
\section{Related Work}

In order to learn a robust captioning system, public datasets have been released for diverse tasks including single image captioning~\cite{lin2014microsoft, plummer2015flickr30k, krishna2017visual}, video captioning~\cite{xu2016msr}, referring expressions~\cite{kazemzadeh2014referitgame, mao2016generation}, and visual question answering~\cite{antol2015vqa, zhu2016visual7w, johnson2017clevr}.
In terms of model progress, recent years witnessed strong research progress in generating natural language sentences to describe visual contents, such as \newcite{vinyals2015show, xu2015show, ranzato2015sequence, anderson2018bottom} in single image captioning, \newcite{venugopalan2015sequence, pan2016hierarchical, pasunuru2017reinforced} in video captioning, \newcite{mao2016generation, liu2017referring, yu2017joint, luo2017comprehension} in referring expressions, \newcite{jain2017creativity, li2018visual, misra2018learning} in visual question generation, and \newcite{andreas2016reasoning, cohn2018pragmatically, luo2018discriminability, vedantam2017context} in other setups. 

Single image captioning is the most relevant problem to the two-images captioning. 
\newcite{vinyals2015show} created a powerful encoder-decoder (i.e., CNN to LSTM) framework in solving the captioning problem. 
\newcite{xu2015show} further equipped it with an attention module to handle the memorylessness of fixed-size vectors.
\newcite{ranzato2015sequence} used reinforcement learning to eliminate exposure bias.
Recently, \newcite{anderson2018bottom} brought the information from object detection system to further boost the performance.

Our model is built based on the attentive encoder-decoder model~\cite{xu2015show}, which is the same choice in \newcite{jhamtani2018learning}.
We apply the RL training with self-critical~\cite{rennie2017self} but do not see significant improvement, possibly because of the relatively small data amount compared to MS COCO.
We also observe that the detection system in \newcite{anderson2018bottom} has a high probability to fail in the three datasets, e.g., the detection system can not detect the small cars and people in spot-the-diff dataset.
The DDLA (Difference Description with Latent Alignment) method proposed in \newcite{jhamtani2018learning} learns the alignment between descriptions and visual differences. It relies on the nature of the particular dataset and thus could not be easily transferred to other dataset where the visual relationship is not obvious.
The two-images captioning could also be considered as a two key-frames video captioning problem, and our sequential multi-heads attention is a modified version of the seq-to-seq model \cite{venugopalan2015sequence}.
Some existing work~\cite{chen2018language, wang2018learning, manjunatha2018learning} also learns how to modify images. These datasets and methods focus on the image colorization and adjustment tasks, while our dataset aims to study the general image editing request task.